\documentclass[10pt,twocolumn,letterpaper]{article}

\usepackage{iccv}              

\definecolor{iccvblue}{rgb}{0.21,0.49,0.74}
\usepackage[pagebackref,breaklinks,colorlinks,allcolors=iccvblue]{hyperref}
\usepackage{multirow}
\usepackage{algorithm}
\usepackage{algpseudocode}
\usepackage{cuted}
\usepackage{todonotes}

\def\paperID{10190} 
\def\confName{ICCV}
\def\confYear{2025}

\title{PaNDaS: Learnable Deformation Modeling with Localized Control}
\author{Thomas Besnier\\
Univ. Lille, CNRS, Centrale Lille, UMR 9189 CRIStAL, F-59000 Lille, France\\
{\tt\small thomas.besnier@univ-lille.fr}
\and
Emery Pierson, Maks Ovsjanikov\\
LIX, Ecole Polytechnique, IPP Paris\\
{\tt\small emery.pierson@lix.polytechnique.fr, maks@lix.polytechnique.fr}
\and
Sylvain Arguillere\\
Laboratoire Paul Painleve, CNRS, UMR 8524, Univ. Lille,  Cité Scientifique, 59655 Villeneuve-d'Ascq\\
{\tt\small sylvain.arguillere@univ-lille.fr}
\and
Mohamed Daoudi\\
Univ. Lille, CNRS, Centrale Lille, Institut Mines-Télécom, UMR 9189 CRIStAL, F-59000 Lille, France\\
IMT Nord Europe, Institut Mines-Télécom, Univ. Lille, Centre for Digital Systems, F-59000 Lille, France\\
{\tt\small mohamed.daoudi@imt-nord-europe.fr}
}

\begin{document}
\maketitle
\begin{strip}
    \centering
    \includegraphics[width=0.9\linewidth]{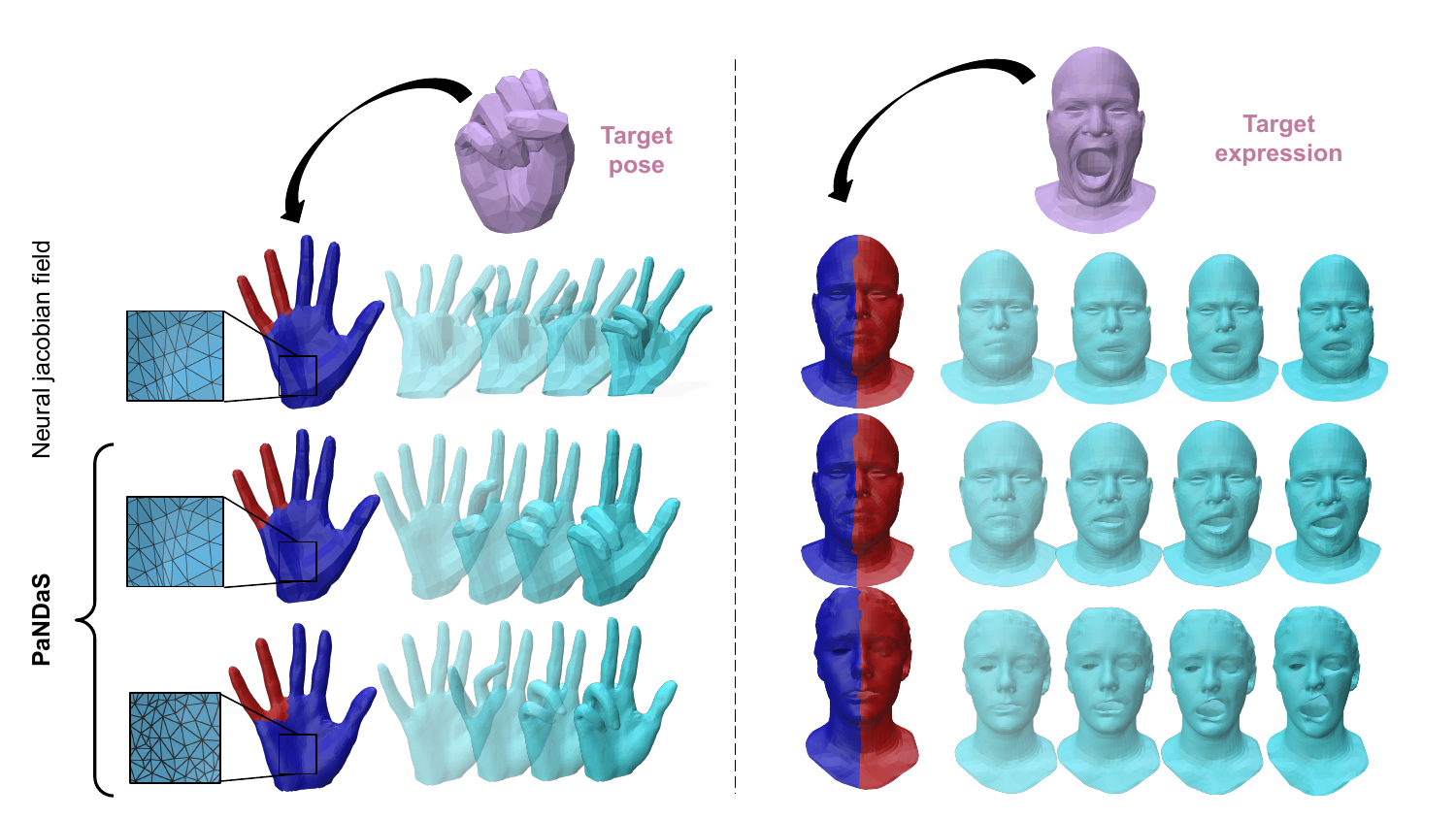}
    \vspace{-3mm}
    \captionof{figure}{We present \textbf{PaNDAS}: a robust learning-based approach to learn non-rigid deformations of triangular meshes applied to human body surfaces. By combining per-face learned features on a neutral pose mesh (in \textcolor{blue}{blue}) with a global encoding of a deformed mesh (in \textcolor{violet}{purple}), the model enables \textbf{localized control} of deformations (in \textcolor{cyan}{cyan}) with user-chosen regions (masked in \textcolor{red}{red} on the left). Notably, this approach permits several applications such as mixing poses, the transfer of partial pose, interpolations and localized shape statistics.}
    \label{fig:LISC_purpose}
\end{strip}

\begin{abstract}

Non-rigid shape deformations pose significant challenges, and most existing methods struggle to handle partial deformations effectively.
We propose to learn deformations at the point level, which allows for localized control of 3D surface meshes, enabling Partial Non-rigid Deformations and interpolations of Surfaces (PaNDaS).
Unlike previous approaches, our method can restrict the deformations to specific parts of the shape in a versatile way. Moreover, one can mix and combine various poses from the database, all while not requiring any optimization at inference time. 
We demonstrate state-of-the-art accuracy and greater locality for shape reconstruction and interpolation compared to approaches relying on global shape representation across various types of human surface data. We also demonstrate several localized shape manipulation tasks and show that our method can generate new shapes by combining different input deformations.
Code and data will be made available after the reviewing process.
\end{abstract}

\section{Introduction}
\label{sec:intro}
\begin{figure}[ht!]
    \centering
    \includegraphics[width=0.8\linewidth]{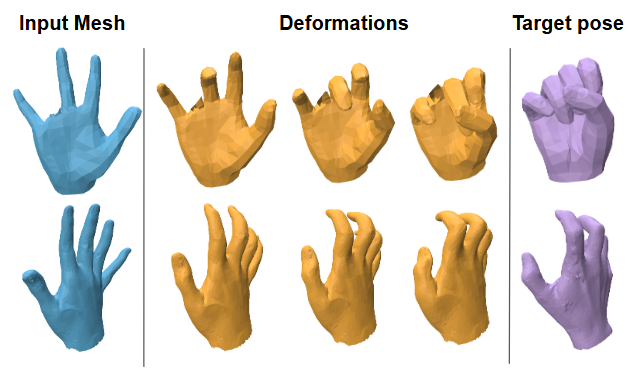}
    \caption{\textbf{Shape interpolation between unregistered hand meshes.} The top row shows a predicted interpolation sequence from a hand with missing ring finger. The bottom row shows an interpolation between two raw scans.}
    \vspace{-3mm}
    \label{fig:infer_non_parametric_hands}
\end{figure}

Generating deformations between 3D human shapes, such as the body, face, and hands, is a cornerstone in human-centered computer vision and graphics, facilitating a broad range of applications like animation, human movement modeling, and character generation. This problem can be generally defined as follows: how to compute a sequence of non-rigid deformations of a source shape so it ends up matching a target shape ? In the case of human body surfaces, something left unsaid is that these deformations and interpolations should correspond to natural human motions to avoid breaking physical plausibility. 
Over the past few years, the problem of computing interpolation between shapes has been addressed extensively. Proposed solutions use geometric regularization~\cite{ARAP_2007, PriMo_2006}, model directly the shape space~\cite{kilian2007geometric, BaRe-ESA_2023_ICCV, HartmanPiersonIJCV2024, ARAP_2000, Bauer_elastic_2011} or, more recently, learn shape interpolation by regularizing neural networks~\cite{LIMP_2020,eisenberger2021neuromorph,SMS_2024}. Most of the approaches rely on a fixed template mesh and do not generalize to unregistered meshes such as in~\Cref{fig:infer_non_parametric_hands}. Neural Jacobian Fields~\cite{NJF_2022} overcomes this problem by predicting the jacobians of the deformations and generalize to new mesh topologies.


\noindent Additionally, this work aims to address a broader version of this deformation problem by also tackling partial deformations as shown in~\cref{fig:LISC_purpose}. More precisely, we focus on computing non-rigid deformations of \textbf{selected parts} of a source shape so it matches the corresponding parts of a target shape while preserving physical plausibility.\\

Most of the cited approaches, including Neural Jacobian Fields, are not designed to address the challenge of partial deformations and interpolations without a complex configuration. Indeed, maintaining geometric consistency and ensuring smooth transitions between deformed and anchored regions usually require manual corrections or intermediate key points~\cite{Shtern2016FastBT}. Moreover, in most of the proposed models, the deformation is extracted from a global latent vector, which makes local control of deformation difficult.

Alternatively, rich and complex shape deformation models have also been constructed through parametric modeling~\cite{FLAME:SiggraphAsia2017, SMPL:2015, MANO_2017}, and such models do enable localized control. However, they typically rely on human input and modeling, making them difficult to extend to arbitrary shape collections. 

Our present work aims to improve the flexibility of the deformation and interpolation processes in a purely data-driven way, to establish a deformation model with localized control on any shape category. 
We propose a novel neural method for Partial Non-rigid Deformations of Surfaces (PaNDaS) to compute local and global near-isometric deformations of triangular meshes.
Instead of representing shapes as a single latent vector, we propose to learn both a global shape latent vector and a local point-wise latent representation. Those vectors are fed into a deformation generator to obtain the desired mesh. Localized learning has been proposed in the specific context of deformation recovery from shape-matching algorithms~\cite{sundararaman2024deformation}, but in this work, we use it to enable local control of the deformation. Indeed, by acting on local point-wise features, our model can interpolate between two shapes on selected local parts by masking the parts one wants to deform.  
As shown in~\cref{fig:LISC_purpose}, PaNDAS can deform specific parts of the whole human body such as fingers on a hand and user-selected parts of faces. 
Additionally, one can apply a mix-and-match strategy to create new poses and compute partial shape statistics, as shown in~\cref{fig:applications}. 
Thanks to the localized control allowed by our local latent deformation, PaNDaS set a new state-of-the-art regarding partial deformations and interpolations of surface meshes, with no additional prior. \\
We summarize the main contributions of this paper as: \emph{(i)}  A new, end-to-end, deep learning framework to generate localized, non-rigid deformations of surface meshes, described in~\cref{sec:approach} and~\cref{subsec:full_def};  
\emph{(ii)} This method is the first to allow vertex-level partial deformations and interpolations without requiring any texture information. This is made possible through the manipulation of predicted deep features as shown by several experiments in~\cref{subsec:partial_def};  
\emph{(iii)} We propose several applications such as new pose generation, partial pose deformation transfer and a robust statistical model of shapes handling partial descriptions. These applications are detailed in~\cref{subsec:applications}.

\section{Related work}
\label{sec:related_work}

\subsection{Geometric methods for shape interpolation}

Shape deformations and shape interpolation are intricately linked as interpolations consist of finding a continuous path of deformations applied to a source shape to match a target shape. To effectively compute these deformations, traditional geometric approaches optimize a deformation energy. The latter can be defined from the well-established As-Rigid-As-Possible (ARAP) \cite{ARAP_2000, ARAP_2007} energy to make deformations close to rotations on a local level, PriMo \cite{PriMo_2006} to ensure local rigidity through a prism formed between pairs of surface triangles and their extrusions. Other approaches directly model shapes as points in a shape space~\cite{LDDMM_2005, Bauer_elastic_2011}, with proper Riemannian metrics, where the deformation corresponds to a geodesic. Several metrics have been proposed, such as the As-Isometric-As-Possible~\cite{kilian2007geometric}, Square Root Normal Fields~\cite{jermyn2012elastic}, or parameterized family of elastic metrics \cite{kurtek2011elastic, Bauer2020ANF, Hartman_H2_elastic_match} with terms penalizing specific deformations such as bending or stretching. In most cases, the induced optimization problem is highly non-convex and can fail dramatically when the initialization is far from the desired solution. To address this performance issue, some methods proposed numerical schemes that simplify the initialization~\cite{laga2017numerical} or reduce the search space~\cite{BaRe-ESA_2023_ICCV}.
However, while some of these approaches could be used for localized manipulations, they often require extensive trial and error to set up the correct constraints for satisfying results. In contrast, we designed our model to deal with both partial and local deformations with limited preprocessing steps.

\subsection{Learning approaches for shape deformation}

More recently, deep learning methods have been used to learn shape deformations. The preferred approach consists of building a latent space using an auto-encoder formulation. Graph convolutional auto-encoders are often used~\cite{gong2019spiralnet++, bouritsas2019neural3dmm, conv_autoencoder_kernel_2020, VAE_mesh_deformations_2018, VAE_localized_2022}, but they tend to overfit information from mesh connectivity. Other approaches, such as 3D-CODED~\cite{groueix2018_3DCODED}, learn auto-encoders for shape matching~\cite{Hahner2024_DISCO, qin2023NFR} but are not optimized for shape deformations and tend to fail when the target deformation is non-linear. Neural Jacobian Fields (NJF)~\cite{NJF_2022} improves the generated deformations by predicting the derivatives of the deformation but does not work well for partial deformations.
To address this problem, other methods propose to regularize the latent space. The authors of LIMP~\cite{LIMP_2020} propose to penalize the deviation of geodesic distances in the latent space. ArapReg~\cite{ARAPREG_2021} penalizes directions in the latent space that affect the ARAP energy, a strategy that has been improved to provide a geometric latent space~\cite{GeoLatent_2023} or to learn correspondences~\cite{yang2024gencorres}. Other methods~\cite{eisenberger2021neuromorph, SMS_2024} directly learn to produce interpolations between pairs of shapes, but they can struggle with details and may fail to generalize to new data.
All these methods focus on learning full shape deformations, which limits applications such as partial shape interpolations. In this paper, we propose a method to overcome this limitation and allow the interpolation of parts of meshes, extending the range of possible deformation sequences by a large margin.

\subsection{Neural methods for Partial Deformations}

While optimization-based algorithms can be adapted to partial deformation, they usually require a custom configuration for each data category. 
A few learning-based methods~\cite{SLIDE, conv_autoencoder_kernel_2020} enable partial deformations by structuring the latent space according to the initial shape structure. This allows the user to modify latent points that affect specific parts of the shape. However, as the relationship between latent points and mesh vertices is predefined before training, the level of detail of partial deformations remains limited. 
Recently, deep methods have started addressing localized control for images~\cite{pan2023draggan, shi2023dragdiffusion}. While extending these to surface data is not straightforward because of the non-euclidean nature of shapes, several works leverage this idea for mesh deformations~\cite{xie2024dragd3drealisticmeshediting, yoo2024apap}, by combining image diffusion models prior with suitable deformation models such as Neural Jacobian fields~\cite{NJF_2022} or by defining key points during movements~\cite{muralikrishnan2024temporalresidualjacobiansrigfree}. However, those methods rely on additional information such as texture maps, user prompts for the diffusion model, or skeletal joints. They also require costly optimization steps, making them impractical for applications in which these priors are unavailable. \\
To overcome this limitation, our method leverages a learned and controllable feature field paired with a deformation generator, which predicts a constrained Jacobian field used to generate partial, natural deformations of the mesh.

\section{Description of our approach}

\begin{figure*}[h!]
    \centering
    \includegraphics[width=0.88\linewidth]{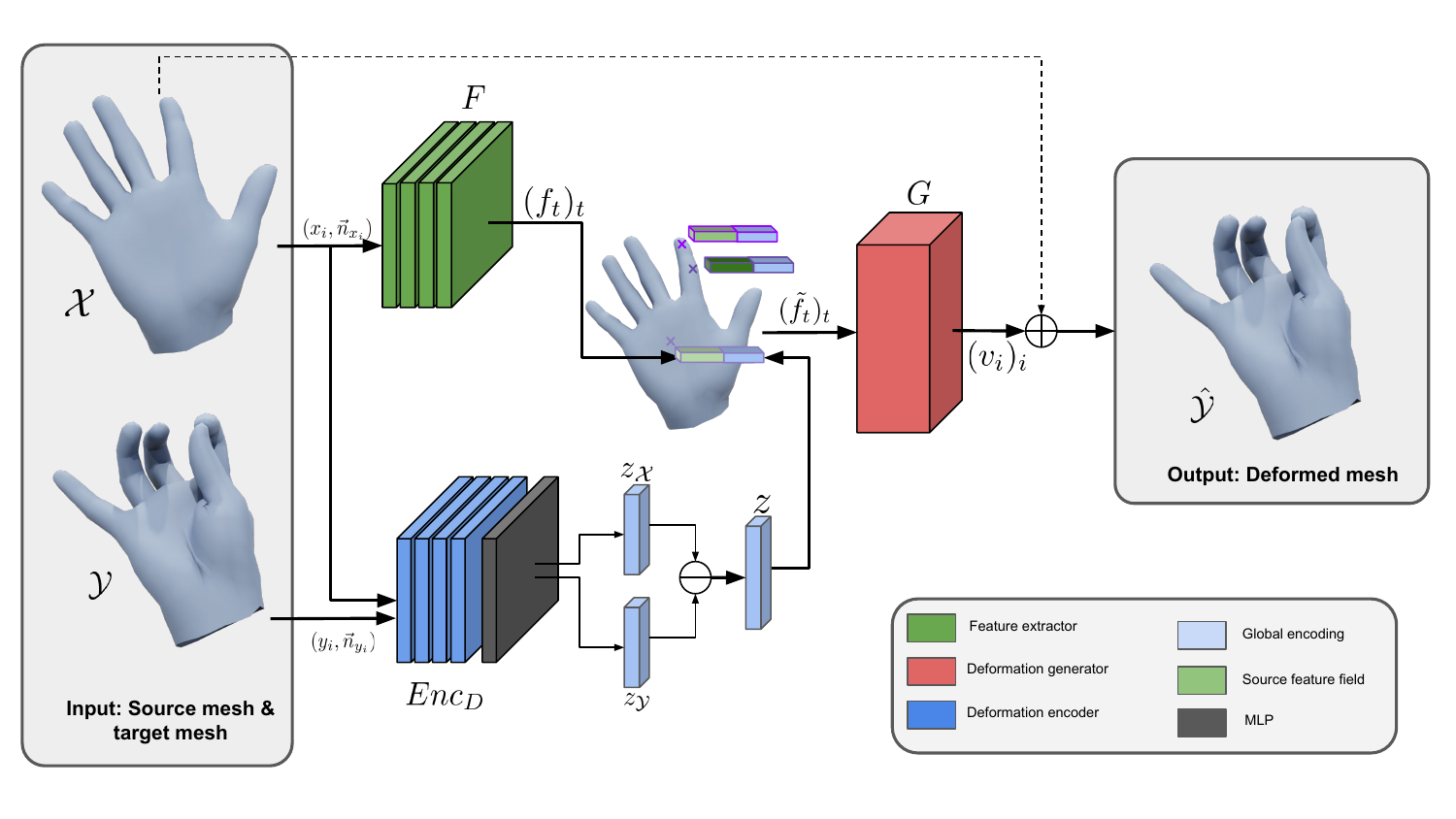}
    \vspace{-5mm}
    \caption{\textbf{Overview of the proposed architecture.} A pair of meshes is given as input: a source pose $\mathcal{X}$ and a target pose $\mathcal{Y}$. Per-triangle features $(f_t)_t$ are learned on the source mesh and are enriched with a global encoding $z$ of the target mesh to obtain a feature field $(\Tilde{f}_t)_t$ over the source mesh. Finally, the deformation generator learns a Jacobian field from the feature field over the neutral mesh to predict a first-order, per-triangle displacement field $(v_i)_{1\leq i \leq n_{\mathcal{X}}}$.}
    \label{fig:global_architecture}
\end{figure*}
\label{sec:approach}
In this section, we explain our approach for which a diagram is associated in~\cref{fig:global_architecture}. More details about the architecture and the training procedure can be found in the supplementary material.

\subsection{Problem formulation}\label{subsec:problem_formulation}
From a source mesh $\mathcal{X} = (x_i)_{1\leq i\leq n_{\mathcal{X}}}$, with $n=n_{\mathcal{X}}$ vertices (the mesh of the ''neutral pose"), we wish to match a target mesh $\mathcal{Y} = (y_i)_{1\leq i\leq n_{\mathcal{Y}}}$, with $n_{\mathcal{Y}}$ vertices.
To do so, we predict a deformation field $v=(v_i)_{1\leq i\leq n}=(v^{\mathcal{X},\mathcal{Y}}_i)_{1\leq i\leq n}$, which morphs $\mathcal{X}$ into $\mathcal{Y}$. 
Our proposed framework for computing $v$ can be divided into 3 parts: a \textbf{neutral pose feature extractor} $F$, a \textbf{deformation encoder} $Enc_D$ and a \textbf{deformation generator} $G$. First, $F$ predicts a local geometric feature vector $f_t \in \mathbb{R}^l$ for each triangle of $\mathcal{X}$, for a total of $l\times m$ features with $m=m_\mathcal{X}$ the number of triangles in $\mathcal{X}$. 
Then, $Enc_D$ encodes both $\mathcal{Y}$ and $\mathcal{X}$ into a single global feature vector $z$. The local (triangle-wise) features $f_t$ and global features $z$ are combined to 
obtain a new local feature vector $\tilde{f}_t=(f_t,z)\in \mathbb{R}^{l+r}$ on each triangle $t$ of $\mathcal{X}$. The deformation generator then computes a deformation field $(v_i)_{1\leq i\leq n}$which is added to the source mesh. 
By expanding the latent code of deformation to the feature field, the model induces locality to the latent code that will allow for local manipulations of the latent code to create new motions and partial interpolations.\\
We now describe in detail how each of those pieces works.

\subsection{Feature field of deformation}

To compute local features on $\mathcal{X}$, we use DiffusionNet \cite{sharp2021diffusionnet} as a face-wise feature extractor $F$ on the source mesh. Heat diffusion is used to define a convolution-like neural network robust to remeshing, with proven efficiency in computing local geometric features~\cite{attaiki2022ncp}. The encoder takes as input the neutral pose mesh $\mathcal{X}$ and outputs a feature vector $f_t\in \mathbb{R}^l$ of size $l=64$ for each triangle $t$ of $\mathcal{X}$. 

Simultaneously, we train a global encoder of deformations $Enc_D$ using DiffusionNet. First, like in the previous section, it predicts a feature vector $g_t$ on each triangle $t$ of $\mathcal{Y}$. Next, we aggregate this entire field of features into a single $r$-dimensional feature vector $z_{\mathcal{Y}}$.
For this, we propose a novel aggregation operator at the end of the network. 
Denote $e^k$ the $k$-th eigenvector of the cotangent Laplacian $\Delta^\mathcal{Y}$ of $\mathcal{Y}$. Note that each $e^k_t$ is a scalar for each face $t$ (remark that $e^1_t$ is constant with respect to $t$, since constant functions belong to the kernel of $\Delta^\mathcal{Y}$). Now $g = (g_t)_t$ can be decomposed as $g_t=p_1e^1_t+p_2e^2_t+\dots$, with each $p_k$ being a $r$-dimensional vector corresponding to the $k-$th frequency of $g$, computed through a weighted average so that
\begin{equation}
    p_k = \frac{1}{\mathrm{Area}(\mathcal{Y})}\sum_{t=1}^{m_\mathcal{Y}} \mathrm{Area}(t)g_t e^k_t.
\end{equation}
Here, $m_\mathcal{Y}$ denotes the number of faces in $\mathcal{Y}$, and $\mathrm{Area}(t)$ is the area of the $t$-th face of $\mathcal{Y}$. In particular, $p_1$ is just the area-weighted average of $g$.
Next, we concatenate the first $s$ frequencies of these projections and pass this vector through a linear layer $Lin$ to obtain a global encoding $z_{\mathcal{Y}} \in \mathbb{R}^r$ of size $r=64$ of $\mathcal{Y}$. In our experiments, $s=4$ proved to be sufficient. Hence, 

\smallskip
\centerline{$
Enc_D(\mathcal{Y})=z_{\mathcal{Y}}=Lin(p_1, p_2, p_3, p_4)
$}
\smallskip

\noindent
is a single vector in $\mathbb{R}^r$. At the same time, this encoder also computes $z_{\mathcal{X}} = Enc_D(\mathcal{X})$ and the final latent vector of deformation is simply $z = z_{\mathcal{Y}} - z_{\mathcal{X}}$. In the particular case $\mathcal{X}=\mathcal{Y}$, we get $z=0$.\\
Now, we go back to $\mathcal{X}$ and $f=F(\mathcal{X})$ from the previous section. For every triangle $t$ of $\mathcal{X}$, append the feature $f_t$ with a copy of the global feature $z$. This gives the final learned feature field $\tilde{f}$, where each face $t$ of $\mathcal{X}$ carries the feature 

\smallskip
\centerline{$\tilde{f}_t=(\tilde{f}_{t,i})_{1\leq i\leq l+r}=(f_t,z)\ \in \mathbb{R}^{l+r}.$}
\smallskip 

\noindent
These features locally describe how a source mesh $\mathcal{X}$ should be deformed to match a target mesh $\mathcal{Y}$. The total number of features is $m_\mathcal{X}\times (l+r)$. \\




\subsection{Deformation generator with Jacobian fields}

From the predicted features put forward by the encoder, the deformation generator uses neural Jacobian Fields \cite{NJF_2022}: it learns per-face candidates for Jacobians matrices $(J_t)_{1 \leq t \leq m_{\mathcal{X}}} \in \mathbb{R}^{3 \times 3}$ of the final deformation field $v$. Differing from the original paper, instead of attaching the triangle centroïds to a global encoding, we predict the Jacobian field directly from the predicted features $(\tilde{f}_t)_{1\leq t\leq m_\mathcal{X}}$. 
Then, these Jacobians are restricted to their corresponding triangle. For notation simplification, we will keep $J_t$ to denote the restricted Jacobian.
Hence, we have $v$ in $\mathbb{R}^{3\times n}$ (recall that $n$ is the number of vertices of $\mathcal{X}$) such that
$J_t = \nabla_t v$
with $\nabla_t$ the gradient operator at triangle $t$.
However, just like any vector field may not be a gradient, any such family of $J_ts$ may not be the Jacobian of a well-defined deformation. This is resolved when taking the closest match for $v$ by solving the Poisson equation 
$\nabla^{\mathcal{X}}v = \nabla^T\mathcal{M}J$, with $\mathcal{M}$ the mass matrix of the source mesh and $J$ the stack of per-face Jacobians $J_t$.
Then, the deformation generator $G$ learns a displacement field $v$ on the template mesh $\mathcal{X}$ so that the model outputs the deformed mesh $\hat{\mathcal{Y}}$ with vertices 
$\hat{y}_i = x_i+v_i$, $i=1,\dots,n$:
\begin{align}
    G : \begin{cases}
        \mathbb{R}^{m_\mathcal{X} \times (l+r)} &\rightarrow \mathbb{R}^{n_\mathcal{X} \times 3}.\\
        (\tilde{f}_{t})_{1 \leq t \leq m } &\mapsto (v_i)_{1\leq i \leq n}
    \end{cases}
\end{align}

\subsection{Training strategy}\label{subsec:training_strategy}

The model is trained end-to-end by minimizing a training loss function $\mathcal{L}$ composed of a reconstruction loss $\mathcal{L}^{rec}$ and an additional term on the normal map $\mathcal{L}^n$
drawn directly from the predicted Jacobians:
\begin{equation}
    \mathcal{L} = \mathcal{L}^{rec} + \lambda^n\mathcal{L}^{n}
\end{equation}
where $\lambda^n$ is a weighting factor.\\
By default, $\mathcal{X}$ and $\mathcal{Y}$ are aligned (there is a point-wise correspondence) but the different component of the model allows changes in the topology of the meshes. With registered data, we use the (Mean Squared Error) MSE for the reconstruction loss:\\
\begin{equation}
    \mathcal{L}^{rec}(\mathcal{Y}, \hat{\mathcal{Y}}) = \frac{1}{n_\mathcal{X}}\sum_{i=1}^{n_\mathcal{X}}\left \|y_i - (x_i + v_i) \right \|_2^2
\end{equation}
Next, we use the normal map to regularize the feature field to obtain smoother deformations. To do so, from the restricted jacobians $J_t$ we extract a normal vector $\Vec{n}_{t,\mathcal{Y}}$ as a cross product.
\begin{figure}[h!]
    \centering
    \includegraphics[trim=90 0 90 0,clip,width=0.9\linewidth]{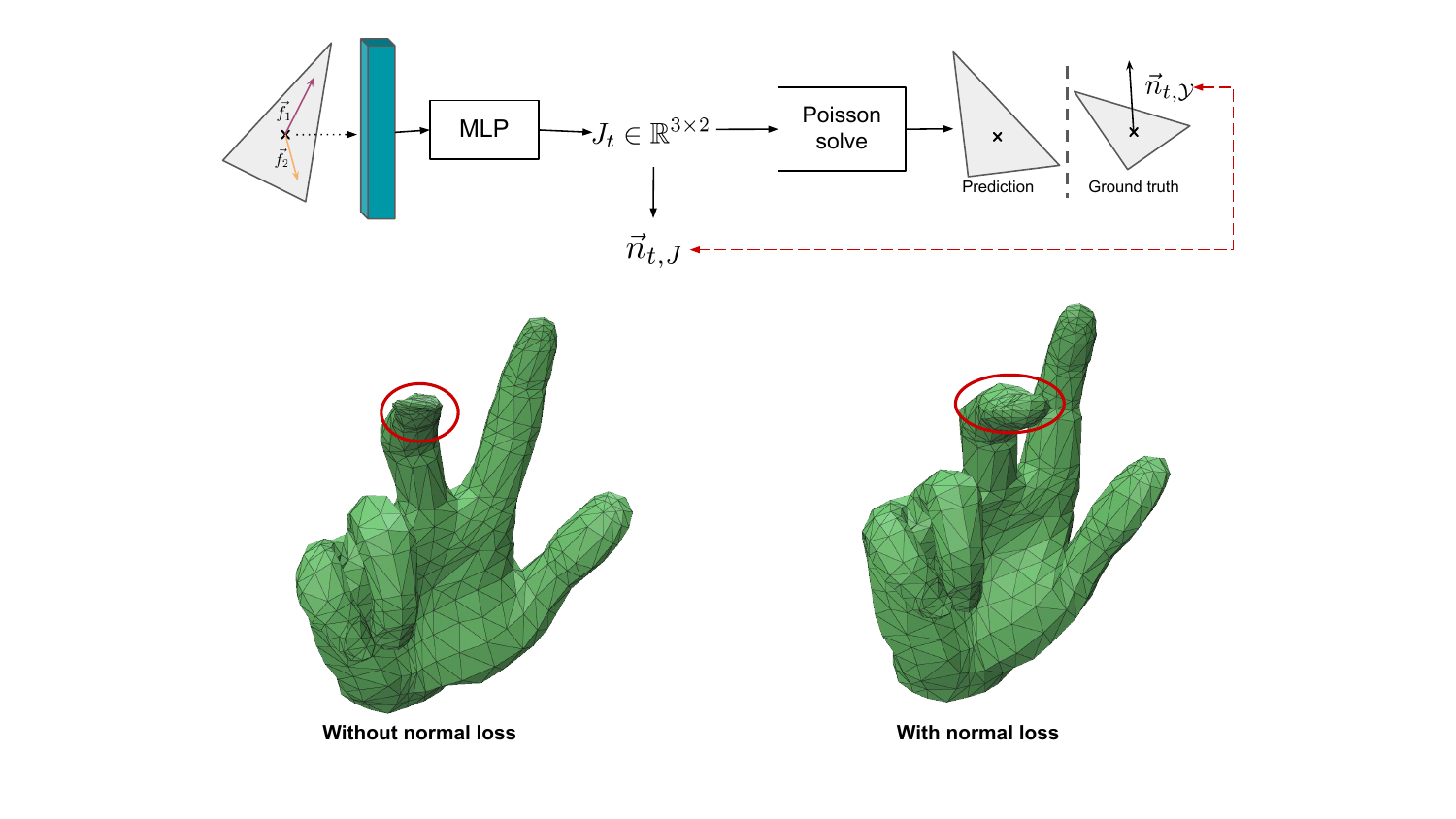}
    \vspace{-5mm}
    \caption{\textbf{Qualitative comparisons of the regularization effect when adding a loss term on the normal map.} On the bottom \textbf{left} is a generation mid-interpolation for a model trained only with the MSE loss, on the \textbf{right} is a similar generated mesh but with a model trained with a loss on the normal map.}
    \label{fig:loss_effect}
\end{figure}
This "intermediate" normal vector will be changed after solving the Poisson equation. However, our experiments showed that using the one from the Jacobian (before the Poisson solve) enhances the quality of deformations from interpolated codes as shown in~\cref{fig:loss_effect}. Then, the loss on the normal map is the Cosine similarity between the target normals $\Vec{n}_{t,J}$ and the predicted normals $\Vec{n}_{t,J}$ at every triangle $t$:\\
\begin{equation}
    \mathcal{L}^n(\mathcal{Y}, \hat{\mathcal{Y}}) = \frac{1}{m_\mathcal{X}} \sum_{t=1}^{m} [1 - \Vec{n}_{t,\mathcal{Y}} \cdot \Vec{n}_{t,J}]
\end{equation}




\subsection{Interpolation, Partial Deformation, Mixing Deformations with masking}
\label{subsec:method_partial}

\begin{figure}[h]
    \centering
    \includegraphics[trim=20 0 20 0,clip,width=1.0\linewidth]{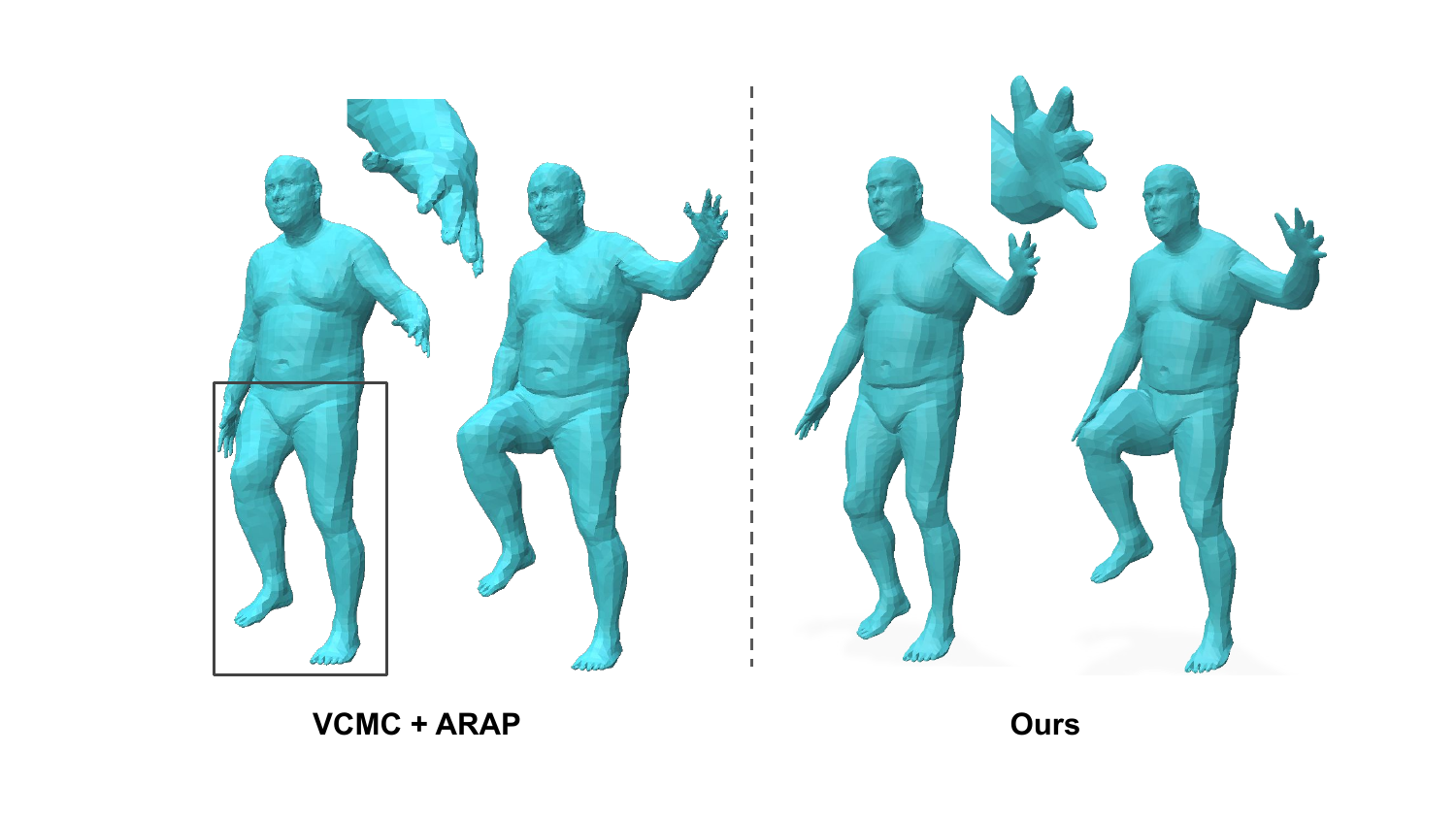}
    \vspace{-10mm}
    \caption{\textbf{Pose mixing using partial interpolations of a neutral pose with two other poses.} The first pose raises both arms, the second pose raises the right knee. We mix them and highlight how our method (on the right) computes smoother deformations and avoids limb shrinking (right leg on the left).}
    \label{fig:mixing_poses_comparisons}
\end{figure}

Once the model is trained, one can perform the usual operations in the latent space. For example, we get an interpolation between $\mathcal{X}$ and $\mathcal{Y}$ by decoding a linear interpolation at each triangle $t$ between $(f_t,0)$ (the encoded features for $\mathcal{X}$) and $\tilde{f}_t=(f_t,z)$ (the encoded features for $\mathcal{Y}$). 
Moreover, we leverage PaNDaS intrinsic locality of the feature field $t\mapsto\tilde{f}_t=(f_t,z)$ to obtain completely new poses. To obtain partial deformations, one can multiply the $z$ part of the local features by a mask $\mathfrak{M}=(\mathfrak{M}_t)_{1\leq t\leq m}\in \{0,1\}^m$, equal to either 1 or 0 on each triangle, to only deform parts of the surface: $\tilde{f}_t^{\mathrm{partial}}=(f_t, \mathfrak{M}_t \odot z)$. This gives more flexibility than previous approaches \cite{conv_autoencoder_kernel_2020} limited to latent points. We can then interpolate between the neutral pose and the partially deformed shape (see~\cref{fig:LISC_purpose}). More generally, we can take several global features $z^1,\dots,z^j$ encoded from several poses $\mathcal{Y}^1,\dots,\mathcal{Y}^j$, assign a part of the shape (and a corresponding mask) to each one and decode the feature field $\left(f_t,\mathfrak{M}^1_t \odot z^1+\dots+\mathfrak{M}^j_t \odot z^j\right)_{1\leq t\leq m}$ to produce new deformations (see~\cref{fig:mixing_poses_comparisons}).


\begin{figure*}[htp!]
    \centering
    \includegraphics[width=0.9\linewidth]{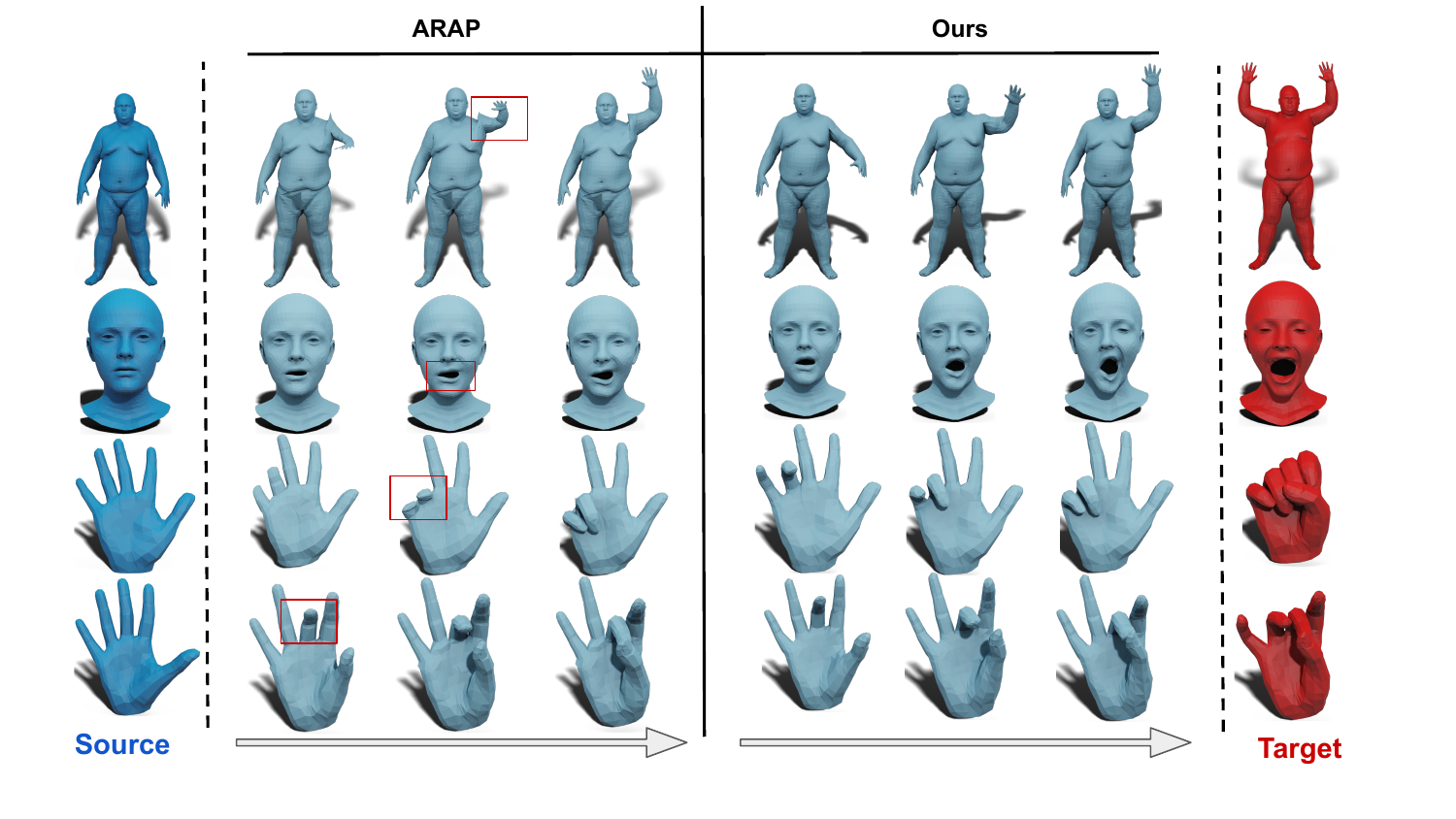}
    \vspace{-8mm}
    \caption{\textbf{Qualitative comparisons of partial deformations on DFAUST, COMA and MANO.} The As-Rigid-As-Possible deformations handle efficiently deformations that are close to linear (2nd and last row) but fail to adapt to large deviations.}
    \label{fig:partial_deformations}
\end{figure*}

\begin{figure}[ht!]
    \centering
    \includegraphics[width=0.9\linewidth]{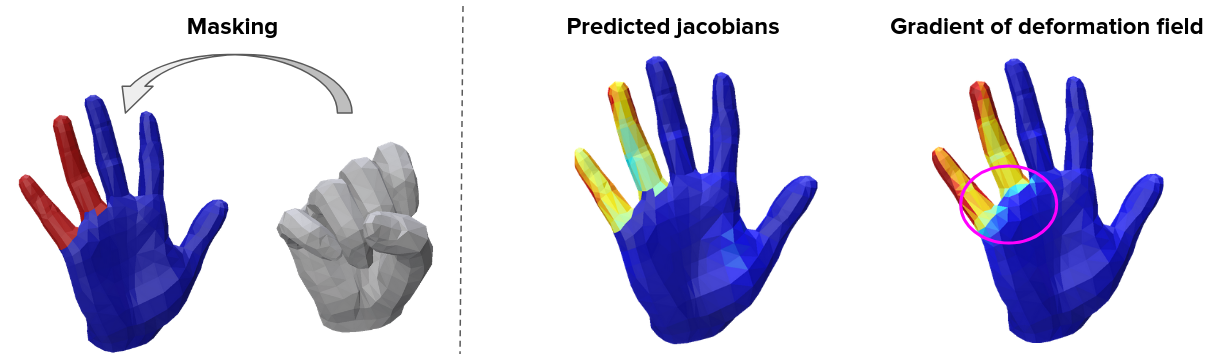}
    \caption{\textbf{Restriction of the deformation field.} We show an example of partial deformation, restricted to the \textcolor{red}{red} parts on the left. We show the corresponding norm of the predicted jacobians and the resulting gradient of the deformation field (after the Poisson solve). We highlight how the deformation is efficiently localized and smoothed around the boundary of the mask (circled in \textcolor{violet}{purple}).}
    \label{fig:norm_masked_comparisons}
\end{figure}

Note that, because of the Poisson solve, there is no guarantee about the deformation being strictly restricted to the masked part. However, we observed that the norm of the deformation decreases rapidly the further away from the mask boundary. We report this observation in~\cref{fig:norm_masked_comparisons} by comparing the norm of the restricted Jacobians to the gradient of the deformation after masking.

\section{Numerical Experiments}
\label{sec:numerical_exp}

\begin{table*}[ht!]
\centering
\footnotesize
\begin{tabular}{@{}l@{\hspace{0.4cm}}ccccccccccc@{}} 
\toprule
& \multicolumn{3}{c}{\textbf{MANO}} & \phantom{abc} & \multicolumn{3}{c}{\textbf{DFAUST}} &
\phantom{abc} & \multicolumn{3}{c}{\textbf{COMA}} 
\\
\cmidrule{2-4} \cmidrule{6-8} \cmidrule{10-12}
& MSE $\downarrow$ & HD $\downarrow$ & CD ($10^{-5}$) $\downarrow$ && MSE $\downarrow$ & HD $\downarrow$ & CD ($10^{-5}$) $\downarrow$ &&  MSE $\downarrow$ & HD $\downarrow$ & CD ($10^{-5}$) $\downarrow$\\
\midrule
LIMP \cite{LIMP_2020} & -  & 0.049 & 71.9 && - & 0.141 & 0.0047 && - & 0.0095 & 1.021 \\
VCMC \cite{conv_autoencoder_kernel_2020} & 0.2240 & 0.014 & 5.86 && 2.92 & 0.076 & 0.0017 && 0.191 & 0.01400 & 1.506 \\
ARAPReg  \cite{ARAPREG_2021} & \underline{0.1986} & \textbf{0.009} & \underline{3.29} && \underline{2.61} & \underline{0.069} & \underline{0.0011} && \textbf{0.128} & \textbf{0.0070} & \textbf{0.474} \\
NJF \cite{NJF_2022} & 0.3019 & 0.019 & 9.65 && 5.01 & 0.126 & 0.0029 && 0.238 & 0.0105 & 1.435 \\
\midrule
\textbf{Ours} & \textbf{0.1422} & \underline{0.010} & \textbf{2.94} && \textbf{2.23} & \textbf{0.058} & \textbf{0.0009} && \underline{0.160} & \underline{0.0086} & \underline{0.839} \\
\bottomrule
\end{tabular}
\caption{\textbf{Reconstruction errors of poses from a given identity.} We evaluate the quality of the deformations using distances between the predicted deformed mesh and the target mesh.}
\label{tab:reconstruction_metrics}
\end{table*}

We showcase the efficiency of our method with three different experiments: the first compares the quality of the generated deformation obtained with the model inference. 
A second experiment compares the \textbf{quality of deformation sequences} obtained through linear interpolation of the feature field. 
Finally, a third experiment shows \textbf{qualitative evaluation of partial interpolations}.\\
Additionally, complex animations performed using the model can be found in the supplementary material videos.

\subsection{Metrics and data}
\textbf{Datasets.} We showcase the adaptability of our method with experiments on 3 different datasets.\\ 
\textit{MANO} \cite{MANO_2017} is a hand mesh dataset comprising 30 hand identities, each in around 20 poses.
We perform a prior Procrustes alignment \cite{procustes_problem} of the meshes.\\
\textit{DFAUST.} \cite{dfaust:CVPR:2017} is a 4D dataset of 10 human subjects that perform 13 different movements. It is made of 40,000 meshes, and we selected 20 meshes per motion from the first 100 frames of each sequence. Our training set is made up of 1140 meshes, and the test set is made up of 300 meshes.\\
\textit{COMA} \cite{COMA_ECCV18} is a face mesh dataset of 12 individuals executing 12 "extreme" expressions. The full dataset contains more than 20,000 meshes, and we uniformly sampled 2000 of them. We also removed the eyeballs that were separated connected components of the face mesh.\\
\textbf{Metrics.} To evaluate mesh reconstructions on registered data, we use the MSE, the Hausdorff distance (HD) and Chamfer distance (CD). Then, we evaluate latent interpolations against ground truth movements. For this, we employ the averaged frame-by-frame MSE and Chamfer distance.

\subsection{Generating deformed mesh}
\label{subsec:full_def}
In this experiment, for each identity in the dataset, we have an identified neutral pose mesh $\mathcal{X}$. Then, given a target mesh $\mathcal{Y}$, the model predicts a deformation field $v$ on $\mathcal{X}$. We observe and measure the quality of the predicted deformed mesh and its distance to the target mesh in~\cref{tab:reconstruction_metrics}. We compared our model with several deep learning methods: Neural Jacobian Fields (NJF) \cite{NJF_2022} proposed an auto-encoding approach to predict a point-wise deformation. Variant Coefficient MeshCNN (VCMC) \cite{conv_autoencoder_kernel_2020} is a graph convolution method, LIMP \cite{LIMP_2020} and ARAPReg \cite{ARAPREG_2021} learn a regularized linear latent shape space of deformations to disentangle identities and poses. 
The localized deformation representation of PaNDaS surpasses global latent interpolation for large non-linear deformation generations (on MANO and DFAUST) and is competitive for face expression deformations. ARAPReg slightly outperforms PaNDaS on the COMA dataset, an easier setting, as most deformations are quasi-linear.


\subsection{Interpolation evaluation}

Secondly, we also validate the quality of latent interpolation sequences using comparisons with short movements in DFAUST and COMA datasets. We report the MSE and mean chamfer distances of the generated sequences in~\cref{tab:interpolation_evaluation_DFAUST} and~\cref{tab:interpolation_evaluation_COMA} against ground truth movements. We compare the performances with latent interpolation from ARAPReg~\cite{ARAPREG_2021} and VCMC~\cite{conv_autoencoder_kernel_2020}.
\begin{table}[h!]
\centering
\scriptsize
\begin{minipage}{0.45\textwidth}
\begin{tabular}{@{}l@{\hspace{0.2cm}}ccc@{\hspace{0.5cm}}ccc@{}} 
\toprule
& \multicolumn{3}{c}{\textbf{MSE $\downarrow$}}  & \multicolumn{3}{c}{\textbf{CD} ($10^{-2}$) $\downarrow$}       \\
& VCMC & ARAPReg  & \textbf{ours} & VCMC & ARAPReg & \textbf{ours} \\
\midrule
punching &   11.98  &  8.94   &  \textbf{8.67}  &  1.02  &  0.85 &  \textbf{0.83}  \\
running &  10.76  &   9.52  &  \textbf{9.40}    &  1.11 & \textbf{1.02} &   1.05  \\
shake a. & 16.48  &  10.98  &  \textbf{10.28}   & 2.56 & 1.73  &  \textbf{1.63}  \\
chicken &  14.82  &  9.17  &    \textbf{9.03}   & 1.21 &  0.90 &  \textbf{0.84}  \\
knees   &  5.02  &  4.52 &   \textbf{4.22}   &  0.52  &  0.38  &  \textbf{0.34}  \\
jumping &  21.46 &  18.08   &    \textbf{17.08}  & 3.74  & 3.18 &  \textbf{2.98}  \\
one leg j. &  6.31  &   \textbf{5.25}  &    5.41    &  0.46 & \textbf{0.38}  &  0.41  \\
one leg l. &  5.57 &   \textbf{4.70}  &    4.94    &  0.43 & 0.40  &  \textbf{0.37}  \\
\midrule
Mean &    11.55   &   8.87  &    \textbf{8.71}   &  1.38  & 1.10  &  \textbf{1.06}   \\
\bottomrule
\end{tabular}
\caption{\textbf{Interpolation evaluation on DFAUST sequences.}}
\label{tab:interpolation_evaluation_DFAUST}
\end{minipage}\hfill
\begin{minipage}{0.45\textwidth}
\begin{tabular}{@{}l@{\hspace{0.2cm}}ccc@{\hspace{0.5cm}}ccc@{}} 
\toprule
& \multicolumn{3}{c}{\textbf{MSE $\downarrow$}}  & \multicolumn{3}{c}{\textbf{CD} ($10^{-5}$) $\downarrow$}       \\
& VCMC & ARAPReg & \textbf{ours} & VCMC & ARAPReg & \textbf{ours} \\
\midrule
bareteeth & 0.16 & 0.12 & \textbf{0.10}     & 1.05 &  0.51 &  \textbf{0.40}   \\
cheeks in & 0.16 & 0.11 & \textbf{0.10}      & 1.08 &  0.38 &  \textbf{0.35}   \\
mouth up & 0.15 & 0.10 & \textbf{0.08}      & 0.93 &  0.41  &  \textbf{0.34}    \\
high smile & 0.15 & \textbf{0.09} & 0.10     &  0.96 & \textbf{0.3}5 &  0.37   \\
lips back & 0.17 & 0.12 & \textbf{0.11}     &  1.15 & 0.44 & \textbf{0.39}    \\
lips up & 0.16 & \textbf{0.08} & 0.09      &  0.92  & \textbf{0.22} & 0.26     \\
mouth side & 0.16 & 0.12 & \textbf{0.11}   &  1.08 &  0.51 &  \textbf{0.43}  \\
mouth ext. & 0.22 & \textbf{0.13} & \textbf{0.13} &  1.46  & 0.52 &  \textbf{0.51}      \\
\midrule
Mean & 0.17  & 0.11 &  \textbf{0.10}     &  1.08  & 0.42 &  \textbf{0.38}     \\
\bottomrule
\end{tabular}
\caption{\textbf{Interpolation evaluation on COMA sequences.}}
\label{tab:interpolation_evaluation_COMA}
\end{minipage}
\end{table}
PaNDaS is competitive compared to these methods on both metrics. To support the quantitative evaluation, we also show qualitative displays of the interpolations in~\cref{fig:compare_interp}. More examples are given in the supplementary material.
It is worth noting that ARAPReg performs worse than on the reconstruction task, primarily because of highly non-linear deformation sequences (see~\cref{fig:compare_interp}).
\begin{figure*}[htbp]
    \centering
    \includegraphics[width=0.78\linewidth]{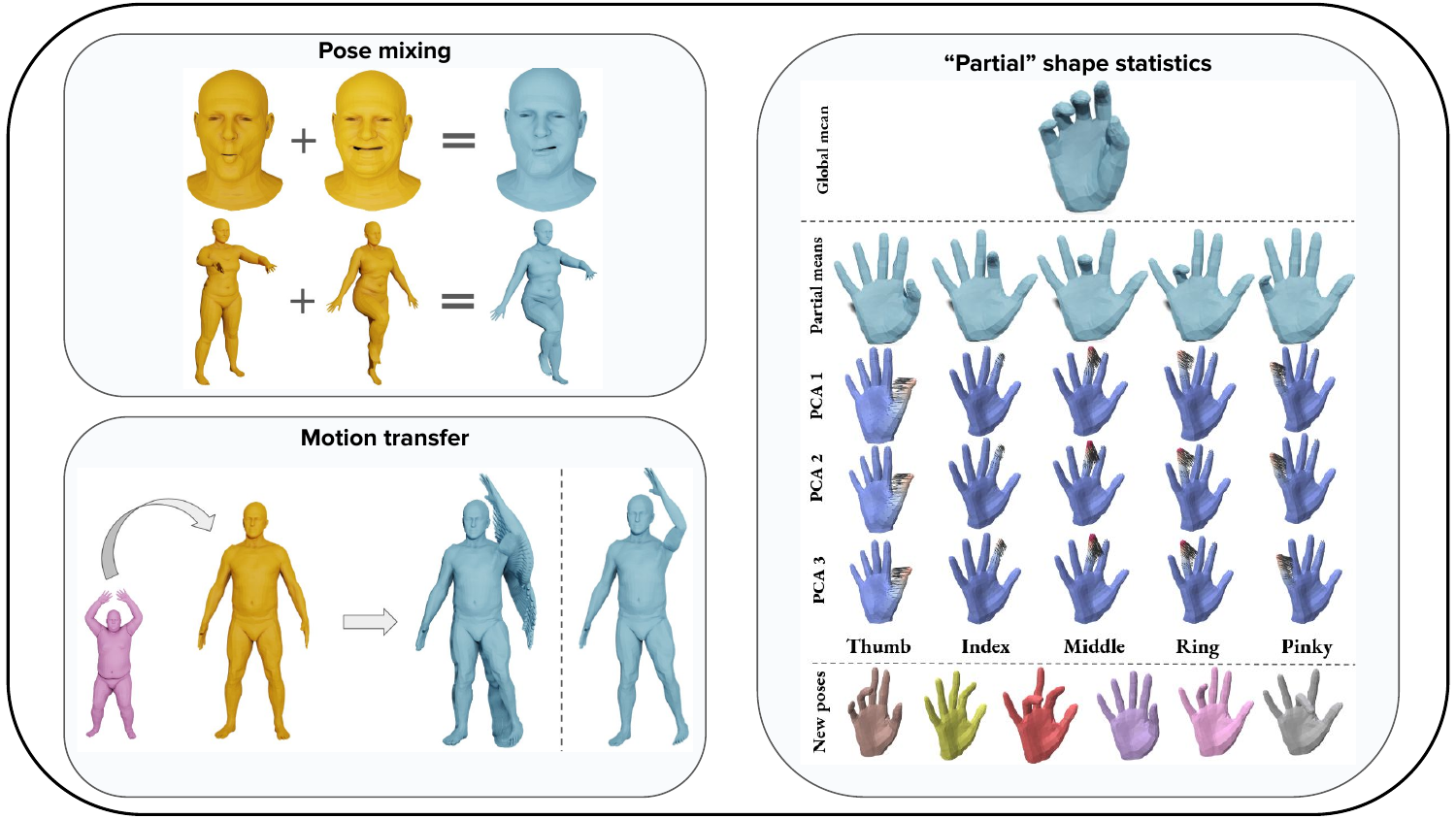}
    \caption{\textbf{Applications of PaNDaS}--\textbf{Pose mixing}: By combining the feature fields of different poses, we can combine deformations to generate new poses. \textbf{Transfer}: The deformation encoder is trained to encode the pose which allow partial motion transfer.  \textbf{Statistics}: The learned feature field is a tensor with euclidean structure on which statistical insights can be mapped back to the space of deformations.}
    \vspace{-5mm}
    \label{fig:applications}
\end{figure*}
\begin{figure}[htbp]
    \centering
    \includegraphics[trim=100 0 100 0,clip,width=0.95\linewidth]{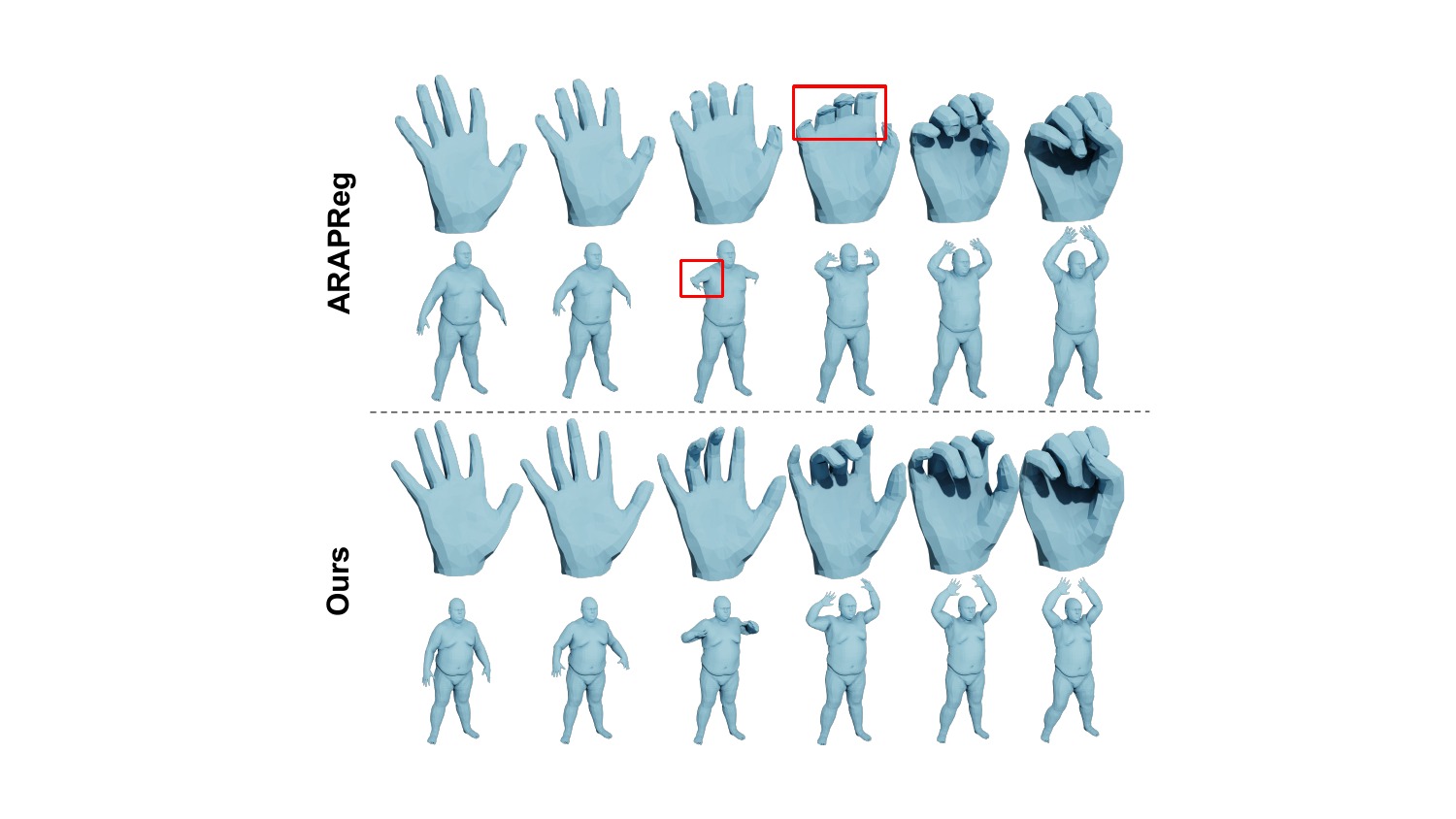}
    \vspace{-8mm}
    \caption{\textbf{Pose interpolation examples.} We compare results obtained using latent linear interpolation with ARAPReg (top row) to the same interpolations computed using PaNDaS (bottom row). The latter handles better large and highly non linear deformations.}
    \label{fig:compare_interp}
    \vspace{-5mm}
\end{figure}

\subsection{Partial deformations and local manipulations}
\label{subsec:partial_def}

For partial deformations, thanks to our training strategy, the generated deformations maintain a global smooth coherent mesh. In particular, it enables the generation of new poses, unseen by the deformation generator. We compare the partial deformations with a localized As-Rigid-As-Possible (ARAP) vertex displacement. The implementation is borrowed from \cite{smooth_corres}. Other methods such as DragD3D \cite{xie2024dragd3drealisticmeshediting} and APAP \cite{yoo2024apap} cannot be used for comparison because no texture map is available in our experiment. The results are presented in~\cref{fig:partial_deformations} and more qualitative examples can be found in the supplementary material.
After intensive hyperparameter tuning, we observe a similar behavior as for full interpolation: for deformations far from a linear interpolation, optimization with ARAP regularization fails to keep local geometric integrity and shrinks the triangles.

\noindent We emphasize the versatility of this method: from a limited collection of different poses, our model enables a wide range of possible partial deformations. Moreover, this deep feature manipulation allows for precise animation of articulated meshes. Notably, it does not require additional skeleton information to produce meaningful movements.

\subsection{Applications: mixing and shape statistics}
\label{subsec:applications}
Our solution offers additional properties that open the way to various applications, as shown in~\cref{fig:applications}.\\
\textbf{Deformation mixing:} As described in \cref{subsec:method_partial}, PaNDaS can mix partial deformations in different parts of the shape. This is achieved using chosen masks on the deep features, which determine how each part is deformed. We can then generate new poses, as illustrated in~\cref{fig:applications}.





\noindent \textbf{A Statistical Model of Deformations:} PaNDaS can be used to estimate the mean shape from a collection of given shapes as well as partial means and principal components, as shown in~\cref{fig:applications}. Interestingly, these components can be used to build a basis for controllable and localized movements.
In addition, because PaNDaS relies on modules that are robust against remeshing, full and partial statistical predictions are possible on unregistered meshes such as scans, for example. More results on this are given in the supplementary material.

\noindent \textbf{Motion transfer.} An interesting property arising from our framework is the ability to transfer motions on selected parts while maintaining the integrity of the whole mesh. Due to a training process focused on learning pose deformations, the deformation encoder naturally disentangle the identity and pose of the target mesh. 



\section{Limitations and future work}
\label{sec:discussion}
Although the PaNDaS architecture is invariant to mesh topology, as other state-of-the-art methods, it is trained with registered meshes. An exciting possibility of future work is to explore recent strategies for unregistered training~\cite{Hahner2024_DISCO, yang2024gencorres}, to train PaNDaS on unregistered data. Moreover, while the quality of partial deformation is satisfying, our simple masking strategy of latent vectors can create local artifacts on the boundary of the selected parts. Improving this strategy, by using weighted masking or learning the mixing of latent vectors, is left for future work.
\section{Conclusion}
In this paper, we presented PaNDaS, a neural method for partial deformations of meshes. Our encoder-decoder architecture, with combined global and local latent vectors, enables the computation of partial deformations of surface meshes. Moreover, our learning strategy allows to learn non-rigid deformations of shapes, and demonstrated state-of-the-art generalization capabilities. We highlighted the efficiency of our method in shape reconstruction, shape global interpolation, and local deformation. Finally, we can easily generate new poses by combining latent codes of training shapes and compute local shape statistics which could make digital avatars more accessible.


\maketitlesupplementary

In this supplementary document, we provide details about the datasets we used, the implementation and hardware specifications of our experiments. Then, we give additional results and comparisons with state-of-the-art methods. Finally, we give more examples and details about potential applications.

\section{Datasets}
The experiments presented were performed on three different datasets. In~\cref{tab:datasets_specs}, we report the data details in terms of quantity (number of sample in the dataset) and resolution (number of points for a given sample). We also report the train/test split. Our model operates on a relatively low data regime.\\
\begin{table}[ht!]
\centering
\begin{tabular}{l@{\hspace{0.5cm}}l@{\hspace{0.3cm}}l@{\hspace{0.3cm}}l@{\hspace{0.3cm}}l}
\toprule
         & \textbf{MANO} & \textbf{DFAUST} & \textbf{COMA}  \\ 
\midrule
Type                 & Hand & Body & Face \\
\# vertices          & 778 & 6890 & 3931  \\
\# faces             & 1538 & 13776 & 7800 \\
Training samples     & 828 & 1140 & 1580  \\ 
Test samples         & 71 & 300 & 336 \\
\bottomrule
\end{tabular}
\caption{\textbf{Train / test splits for each dataset.}}
\label{tab:datasets_specs}
\end{table}\\
Since our framework takes positions and normals as input, all data are rigidly aligned before training and inference. For DFAUST and COMA, the data is already aligned. For MANO, we perform a Procrustes alignment of the training data (this is done for all compared methods).

\section{Implementation and hardware details}
All PaNDaS models presented in the main paper are trained using 1000 epochs using ADAM \cite{ADAM} optimizer with a learning rate of $10^{-4}$. In the experiments, for the loss function, the weighting factor $\lambda^n$ is set as $\lambda^n = 10^{-5}$.\\
All models training and inferences were performed with a computer running Linux with an Intel Xeon Gold 5218R CPU with 64Go RAM and a NVIDIA Quadro RTX 6000 graphic card with 24Go GPU RAM.


\section{Ablation Study}
\label{sec:ablation}
The following ablation study on the MANO dataset highlights the importance of each of PaNDaS' components, namely the encoder, decoder, and loss functions. We evaluate the final MSE on the test set after changing the different components. We report all results in~\cref{tab:ablation_studies}.
Notably, we observe that DiffusionNet is crucial for the encoding part. Moreover, we observe a large drop in performance without our global aggregation strategy. Similarly, a simple MLP fails to decode properly the shapes. Finally, we tested other loss functions, the simple MSE and using ARAP energy, similarly as in~\cite{ARAPREG_2021}. Notably, adding an ARAP regularization term did not improve either the quality of the deformation nor the interpolations quality.

\begin{table}[ht!]
\centering
\begin{tabular}{lll}
\toprule 
 &     & \textbf{MSE} $\downarrow$ \\
\midrule
\multirow{1}{*}{\textit{Encoder}} & MLP  &  0.2456   \\
\midrule
\multirow{1}{*}{\textit{Aggregation}}  & Mean &  0.2286   \\
\midrule
\multirow{2}{*}{\textit{Decoder}} & MLP  &  0.2216   \\
                         & DiffusionNet           &  0.2074   \\
\midrule
\multirow{2}{*}{\textit{Loss}}    & $\mathcal{L}^{rec}$  &  0.1610   \\
                         & $\mathcal{L}^{rec}$ + ARAP  &  0.1790  \\
\midrule
                        & \textbf{Ours}  &  \textbf{0.1422} \\
\bottomrule
\end{tabular}
\caption{\textbf{Ablation studies on the MANO dataset.} The overall approach of the framework gives satisfying results with simpler components. However, our best results were obtained by combining DiffusionNet for the feature extractor, aggregation by projecting on the eigenvectors, deformations generated with Jacobian fields and loss defined as $\mathcal{L} = \mathcal{L}^{MSE} + \lambda_n\mathcal{L}^n$.}
\label{tab:ablation_studies}
\end{table}

\section{Additional qualitative examples}
We provide more qualitative results and comparisons of latent interpolations in this section.

\subsection*{Full deformations}
We first display examples of full deformations generated from latent interpolations in~\cref{fig:compare_interp}, and a comparison against ARAPReg.

\begin{figure}[h!]
    \centering
    \includegraphics[width=\linewidth]{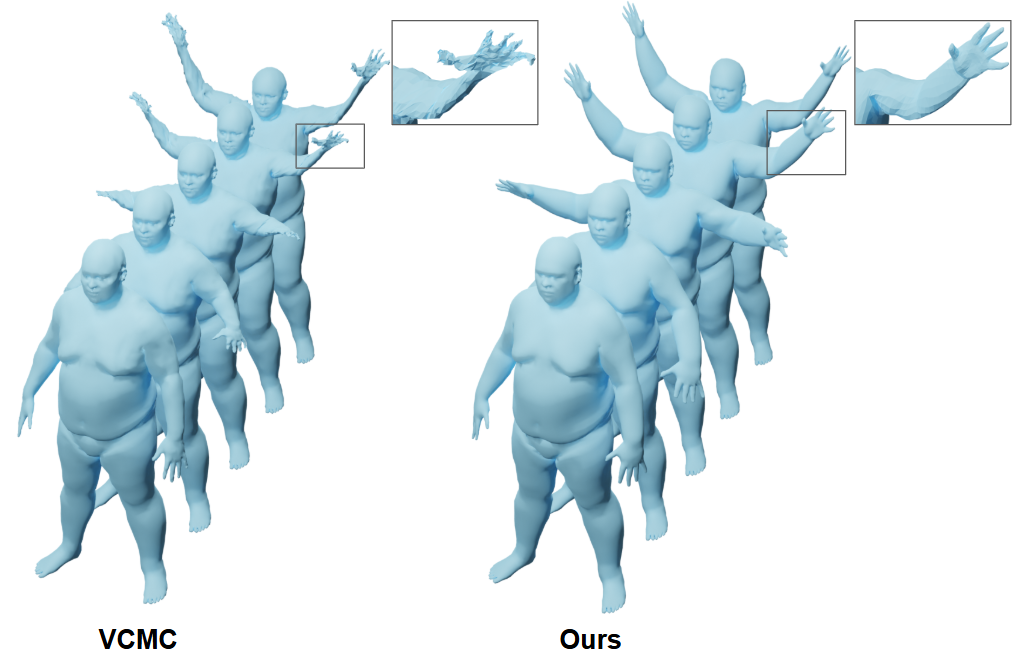}
    \caption{\textbf{Qualitative comparison of latent interpolation.} Our method shows more realistic interpolation trajectories, with limited distortion of the hand even with large non-linear deformations.}
    \label{fig:compare_interp}
\end{figure}

\subsection*{Partial deformations}
\begin{figure}[h!]
    \centering
    \includegraphics[width=\linewidth]{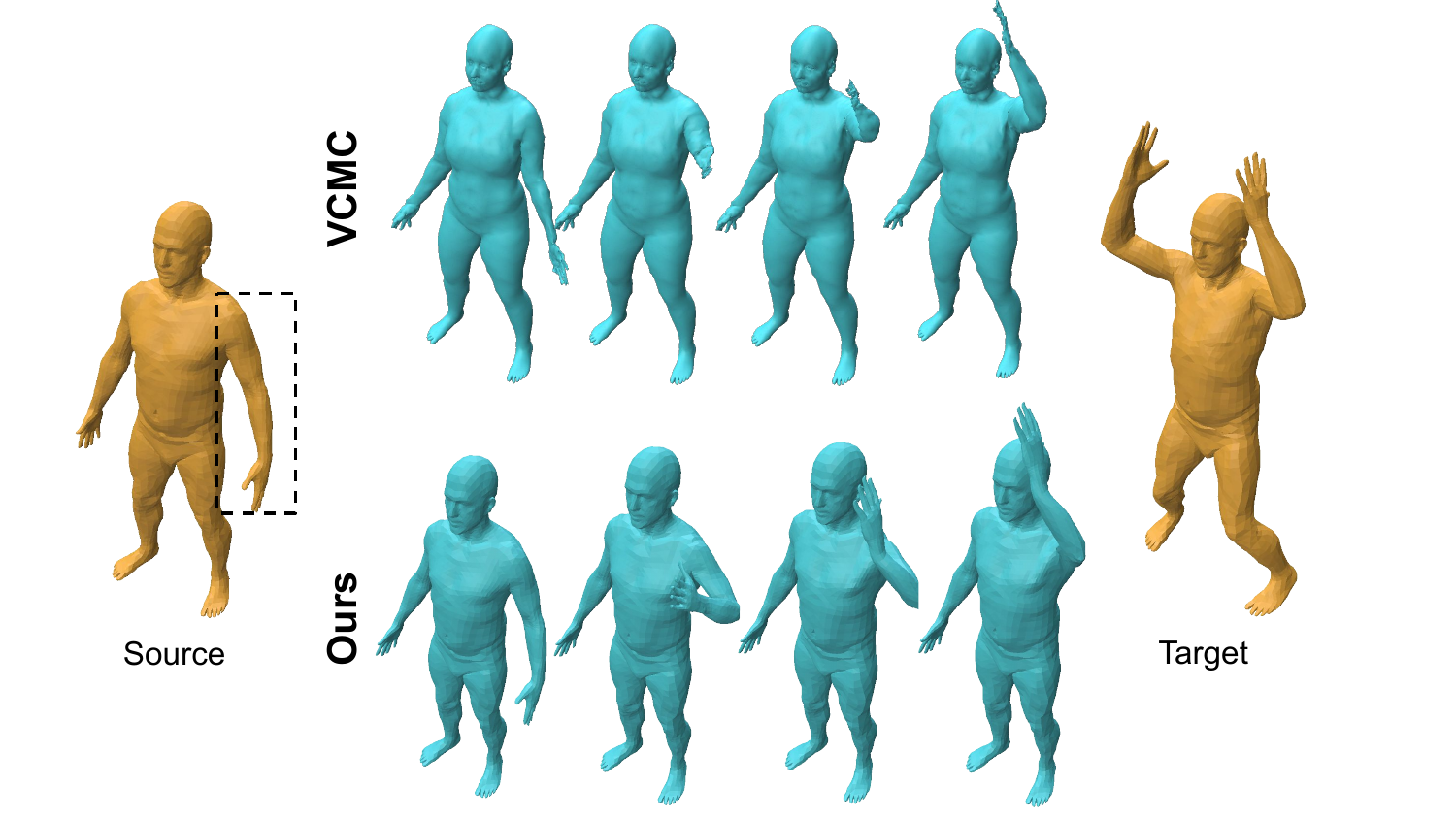}
    \caption{\textbf{Qualitative comparison of partial latent interpolation.} We computed a partial latent interpolation on an unseen identity between a \textit{source} pose and a \textit{target} pose displayed in yellow. The interpolated part is the left arm (dashed case). Previous comparable method VCMC fails to reconstruct the identity and exhibit important distortions of the selected part during the interpolation.}
    \label{fig:compare_partial_VCMC}
\end{figure}

Then we give another comparison of partial shape interpolation in~\cref{fig:compare_partial_VCMC}.

\begin{figure}[h!]
    \centering
    \includegraphics[width=\linewidth]{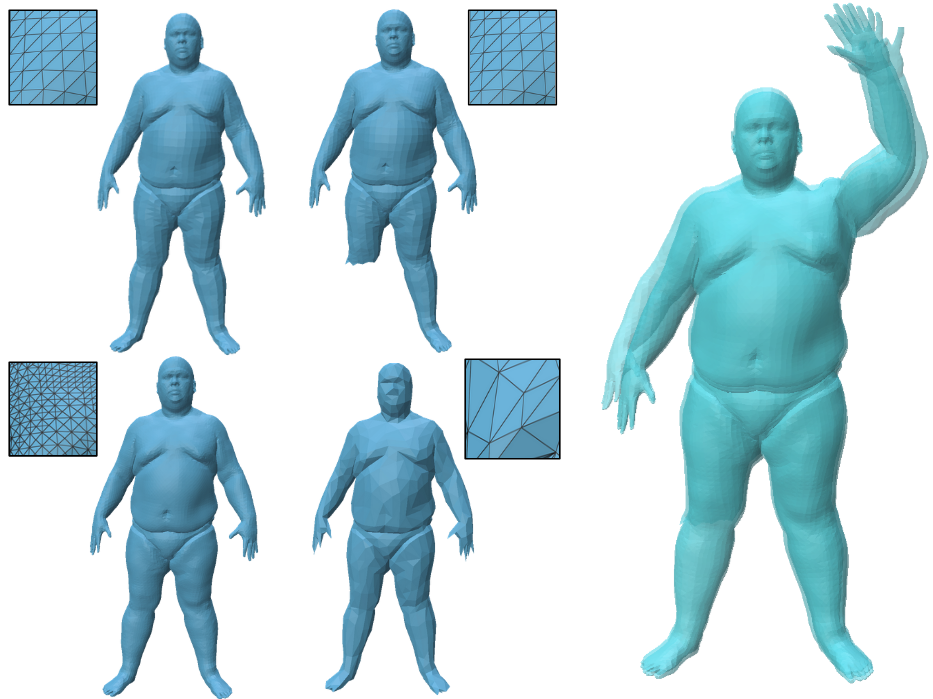}
    \caption{\textbf{Robustness of partial deformations.} From different meshing of the source mesh (on the left), we show that the resulting partial deformations restricted to the left arm stays similar by superposing the results (on the right).}
    \label{fig:robustness_DFAUST_exp}
\end{figure}

\noindent Thanks to the different component of the framework, restricting deformations to specific is robust against remeshing as shown in~\cref{fig:robustness_DFAUST_exp}.

\subsection*{Applications: Shape statistics}
\begin{figure*}[h!]
    \centering
    \includegraphics[width=\linewidth]{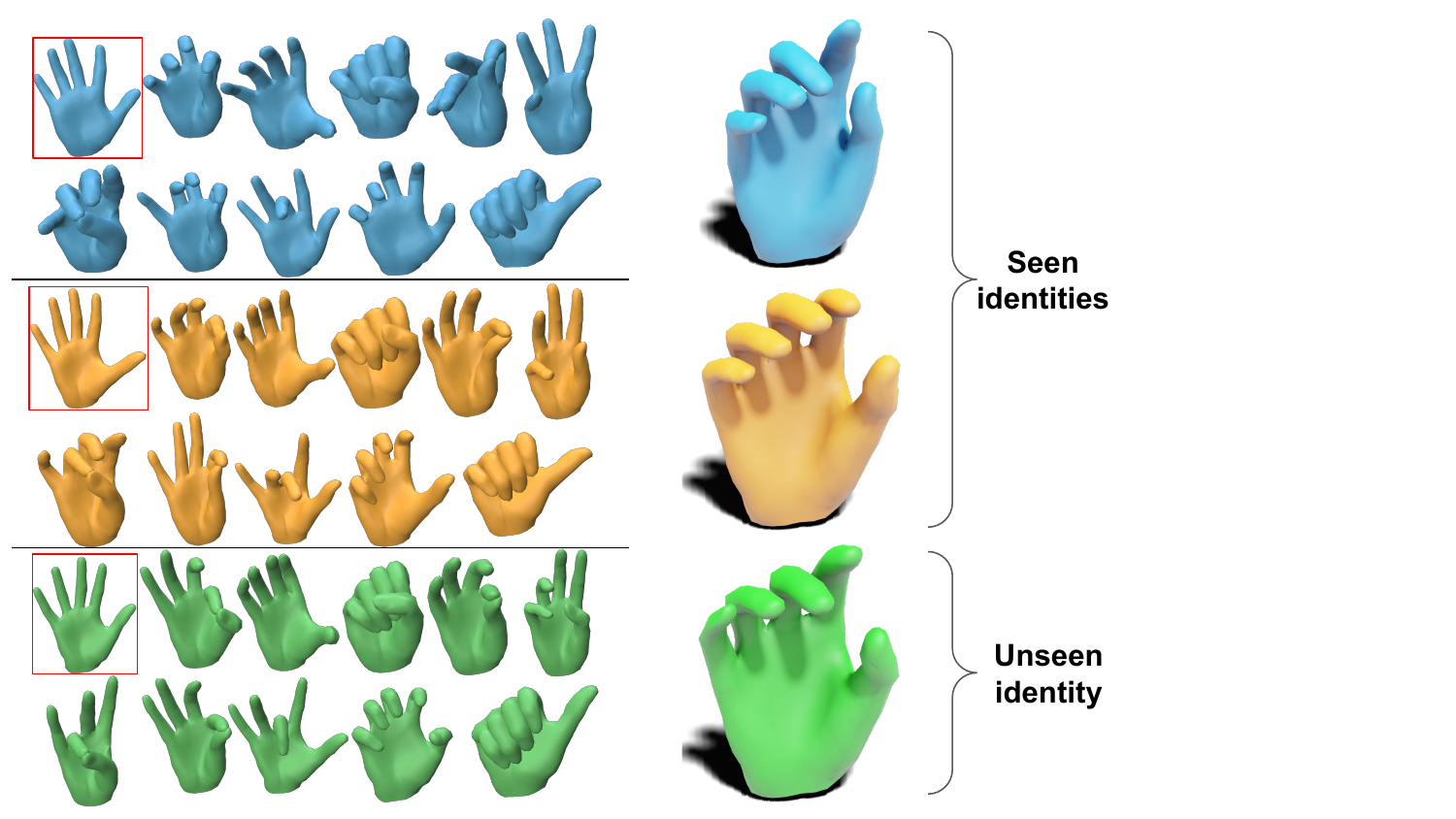}
    \caption{\textbf{Robustness of the mean shape estimation.} We perform mean shape estimation from a collection of 11 poses of hands. We predict the feature field associated with each pose and compute the Euclidean mean. The mean feature field is then used to compute the \textbf{mean pose}. In this figure, each color corresponds to a different identity (framed in red on the left). In particular, we highlight the robustness of our method across several hand identities, seen and unseen during training.}
    \label{fig:mean_robustness}
\end{figure*}

As mentioned in the main paper, PaNDaS can be used to build statistical models of non-rigid deformations, such as the mean of shape collections. 
We demonstrate the robustness of our method by showing in~\cref{fig:mean_robustness} the generalizability of our trained model. The estimated mean from similar poses using PaNDaS is consistent across different identities.

Additionally, we show results and comparison of mean estimation in

\begin{figure}[h!]
    \centering
    \includegraphics[width=1.0\linewidth]{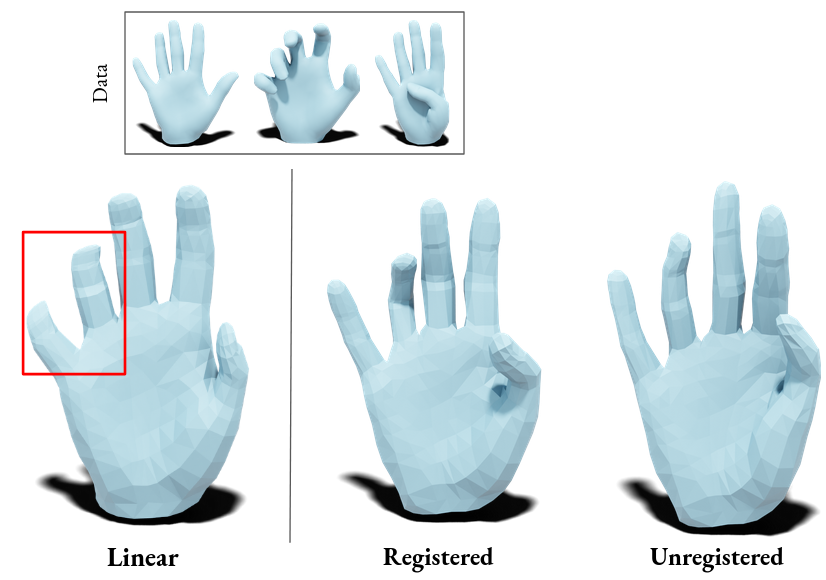}
    \caption{\textbf{Estimated mean from the same collection of hands.} From 10 different hand pose from MANO, we compute a linear mean (on the left) and the "latent mean" of the registered meshes along with the latent mean of the raw scans (unregistered).}
    \label{fig:mean_comparison}
\end{figure}

\subsection*{Applications: Partial Expression transfer}
\begin{figure}[h!]
    \centering
    \includegraphics[width=1.2\linewidth]{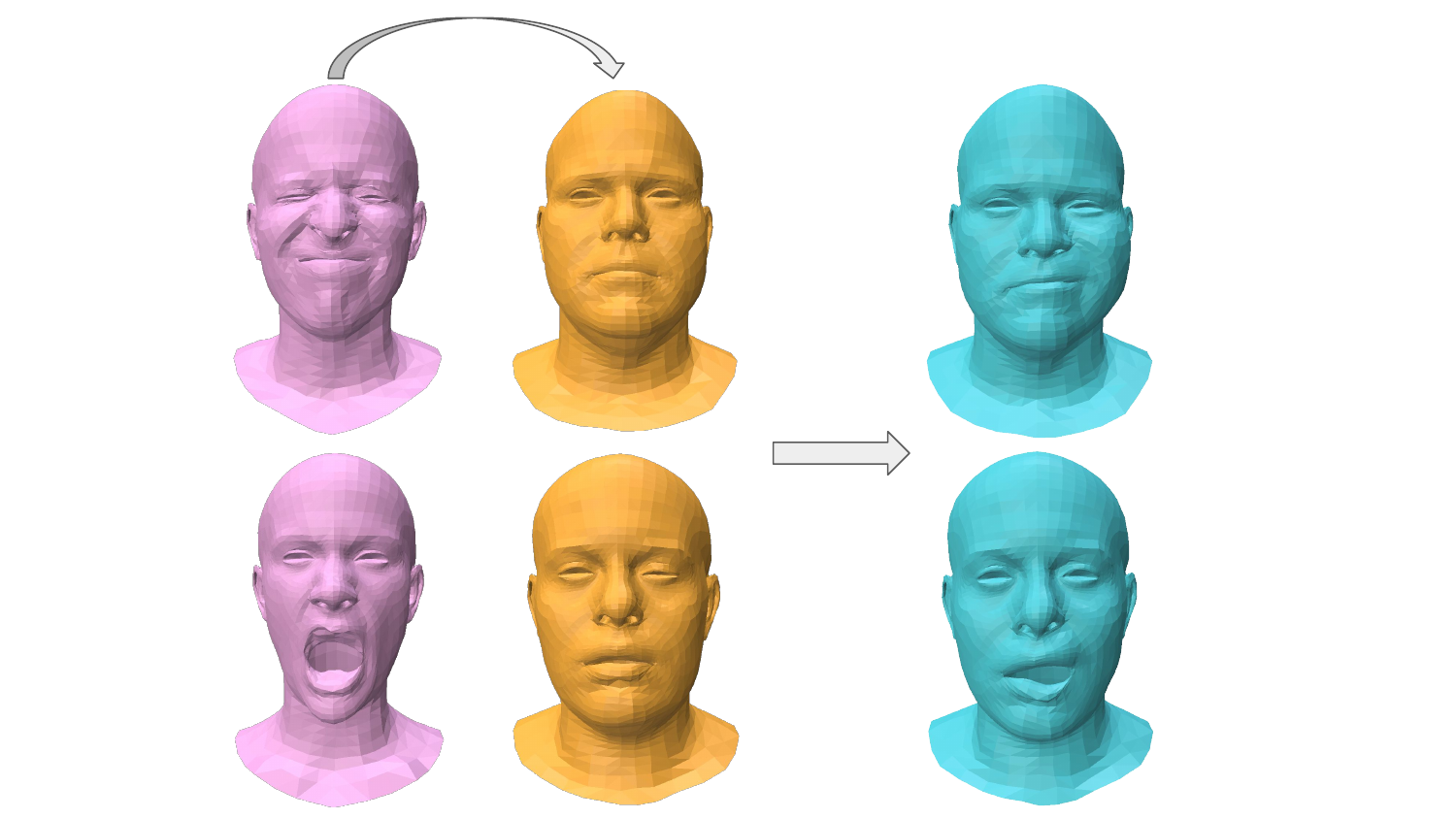}
    \caption{\textbf{Example of partial expression transfer.} In this example, the expression of the face in red is transferred to the yellow face restricted on the left side.}
    \label{fig:partial_expr_transfer}
\end{figure}

We give examples of partial expression transfer in~\cref{fig:partial_expr_transfer}.

\section{Animations}
We attached to the supplementary material, several videos showing animations (on the left side of the videos) computed with partial deformations extracted from combinations of different input poses (on the right side of the videos). PaNDaS can generate complex, controllable motions from a limited number of input poses.

{

\clearpage
    \small

}

\end{document}